# Reading Between the Lines: How Electronic Nonverbal Cues shape Emotion Decoding

**Taara Kumar**[1]
**Kokil Jaidka**[1,2]

[1]Department of Communications and New Media,
[2]Centre for Trusted Internet and Community,
National University of Singapore, Singapore

## Abstract

As text-based computer-mediated communication (CMC) increasingly structures everyday interaction, a central question re-emerges with new urgency: How do users reconstruct nonverbal expression in environments where embodied cues are absent? This paper provides a systematic, theory-driven account of electronic nonverbal cues (eNVCs)—textual analogues of kinesics, vocalics, and paralinguistics—in public microblog communication. Across three complementary studies, we advance conceptual, empirical, and methodological contributions. Study 1 develops a unified taxonomy of eNVCs grounded in foundational nonverbal communication theory and introduces a scalable Python toolkit for their automated detection. Study 2, a within-subject survey experiment, offers controlled causal evidence that eNVCs substantially improve emotional decoding accuracy and lower perceived ambiguity, while also identifying boundary conditions, such as sarcasm, under which these benefits weaken or disappear. Study 3, through focus group discussions, reveals the interpretive strategies users employ when reasoning about digital prosody, including drawing meaning from the absence of expected cues and defaulting toward negative interpretations in ambiguous contexts. Together, these studies establish eNVCs as a coherent and measurable class of digital behaviors, refine theoretical accounts of cue richness and interpretive effort, and provide practical tools for affective computing, user modeling, and emotion-aware interface design. The eNVC detection toolkit is available as a Python and R package at https://github.com/kokiljaidka/envc.

## Introduction

As reliance on computer-mediated communication (CMC) grows—through text messaging, emails, and social media, the question of how people understand one another without access to embodied cues has renewed significance. Facial expressions, gestures, and vocal tone convey affective nuance in face-to-face settings (Burgoon, Manusov, and Guerrero 2022). In their absence, readers reconstruct emotional meaning from whatever surface features remain, often in length-constrained mi-

---

croblog posts, on platforms such as X (formerly Twitter). These reconstructions draw not only on added cues but also on the absence of expected cues. For instance, flat punctuation, minimal emphasis, or stylistic uniformity may themselves be interpreted affectively. From the outset, digital emotion understanding involves a tension between what is present and what is missing, and subjective clarity does not always align with author intent.

A primary resource enabling reconstruction is digital prosody: orthographic and stylistic devices that approximate vocal tone in written form. Elongation, expressive punctuation, repeated letters, spacing conventions, and laughter or vocalization markers allow writers to gesture toward intensity, stance, or rhythm. Prior work demonstrates that these cues meaningfully shape affective interpretation (Rodríguez-Hidalgo, Tan, and Verlegh 2017; Hancock 2004; Walther and Tidwell 1995). Yet in practice, their effects are far from uniform. This inconsistency foregrounds a deeper theoretical puzzle that has remained largely unaddressed. In speech, prosodic cues reliably enhance affective clarity. Online, however, the same surface marker can either clarify emotional intent or complicate it. An elongated vowel may amplify genuine enthusiasm or mark ironic exaggeration; a smiling emoji may convey warmth or function as a sarcasm cue; ellipses can soften tone or signal irritation. Thus, the functional load carried by eNVCs depends heavily on message characteristics and genre norms, not simply on the presence of a cue.

This variability creates a empirical gap, where media richness perspectives (Daft and Lengel 1986) suggest that adding expressive resources increases clarity, whereas Electronic Propinquity Theory (Korzenny 1978; Korzenny and Bauer 1981) highlights that additional signals can increase interpretive effort, especially when cues misalign with textual content. Both perspectives, however, imply that the effects of cues are conditional rather than universal. The central theoretical problem is therefore not whether eNVCs facilitate or block reader interpretation, but rather when and why their interpretive value changes.

Sarcasm exemplifies this boundary condition, as many of the same eNVCs used to intensify literal emo-tion are used to mark its inversion (Thompson and Filik 2016; Hancock 2004; Danesi 2016). However, despite extensive research on sarcasm in dyadic messaging, no prior controlled study has jointly manipulated sarcasm and eNVC presence in naturalistic microblog content using a within-subject design. Identifying these boundary conditions is essential for refining theories of cue richness and for grounding EPT's claims about interpretive effort in contemporary communication practices.

Motivated by these gaps, we examine three research questions:

- **RQ1: How do nonverbal cues present themselves in microblog posts?**
- **RQ2a: Do eNVCs help readers accurately infer author-intended emotions in microblog posts?**
- **RQ2b: Does sarcasm moderate the effectiveness of eNVCs in supporting emotional interpretation?**
- **RQ3: How do people interpret emotions through eNVCs in microblog posts?**

We note at the outset that our taxonomy focuses on orthographic and symbolic cues: the kinesics and paralinguistics rendered in text. Other modalities prevalent on social media, such as GIFs, memes, video, and hashtag pragmatics, also carry nonverbal meaning but fall outside the scope of text-only analysis. We return to this boundary in the Discussion.

To answer RQ1, we develop a taxonomy aligned with foundational nonverbal communication theory, focusing on kinesic equivalents (textualized facial or bodily expressions) and paralinguistic equivalents (prosodic approximations of loudness, rhythm, and vocalization) (Birdwhistell 1952; Burgoon, Manusov, and Guerrero 2021; Hall, Horgan, and Murphy 2019). We apply this taxonomy to a large Vent corpus using a transparent regex-based toolkit. To answer RQ2, we introduce a within-subject experiment that systematically manipulates eNVC presence and sarcasm. By balancing stimuli across a 2×2 design, Study 2 allows us to identify when eNVCs facilitate interpretation and when they are neutralized by cue–text misalignment. To answer RQ3, focus-group discussions reveal the interpretive strategies people use in situ, including how they integrate cues with contextual expectations and how the absence of expected cues becomes meaningful.

Across the three studies, we show that eNVCs reliably enhance emotional interpretation for literal posts but that their benefits attenuate in sarcastic or contradictory contexts. The results reveal a dissociation under sarcasm, where uncertainty rises and accuracy falls regardless of cue presence—showing that eNVCs cannot restore either subjective clarity or correct inference when affect is ironic. This project thereby combines computational, experimental, and qualitative methods to delineate the key boundary conditions under which digital nonverbal cues facilitate or complicate affective understanding in contemporary text-based communication.

## Background

### Nonverbal Communication

Nonverbal communication refers to "the transfer and exchange of messages in any and all modalities that do not involve words" (Matsumoto, Frank, and Hwang 2013, p. 4). Classic taxonomies distinguish appearance, environment, and dynamic expressive channels such as facial movement, gesture, posture, and vocal tone (Ekman and Friesen 1969; Hall and Knapp 2013; Burgoon, Manusov, and Guerrero 2021). Core systems include kinesics, oculesics, haptics, proxemics, chronemics, and paralinguistics (Argyle 1975; Birdwhistell 1952; Hall, Horgan, and Murphy 2019).

Foundational estimates of verbal–nonverbal influence (Mehrabian and Wiener 1967; Mehrabian and Ferris 1967) were limited to specific lab tasks and have since been reinterpreted (Burgoon, Manusov, and Guerrero 2021; Knapp, Hall, and Horgan 1972). Subsequent work demonstrates that prosody conveys emotional and pragmatic meaning independent of lexical content. Studies of prominence, pitch movement, rhythm, and temporal structure show how speakers use gradient loudness (Kochanski et al. 2005; Cole 2015), pitch accent (Watson, Tanenhaus, and Gunlogson 2008), and intonational contour (Bolinger and Bolinger 1986; Ladd 2008) to signal stance and affect. Prosodic modulation is therefore a central carrier of emotional meaning (Scherer 2003; Ogden 2017).

In text-only settings these acoustic resources are absent, and users adopt orthographic strategies—elongation, expressive punctuation, spacing conventions, vocalization markers—as forms of digital prosody (Herring and Androutsopoulos 2015; McCulloch 2019; Crystal 2006). Existing work has primarily focused on affective cues such as emojis, emoticons, and GIFs (Derks, Fischer, and Bos 2008; Jiang, Fiesler, and Brubaker 2018; Waterloo et al. 2018; Price and Robinson 2021), with less attention to specific paralinguistic conventions like elongation, repetition, or vocalization markers (Provine, Spencer, and Mandell 2007; Krishnan and Hunt 2019; Tagliamonte 2016; Baron 2002). These studies are fragmented across fields and often examine isolated cue types, limiting integration with foundational nonverbal theory.

We address this gap by grounding electronic nonverbal cues (eNVCs) in canonical nonverbal systems. Our taxonomy identifies (1) kinesic equivalents (textual gestures, facial or bodily emojis) and (2) paralinguistic equivalents (orthographic loudness, pitch, rhythm, vocalizations). This mapping links digital prosody to theories of expressive behavior and supports systematic, large-scale analysis. Given the absence of a unified framework mapping classical NVC categories to digital text, we ask:

**RQ1: How do nonverbal cues present themselves in microblog posts?**

### Theory of Electronic Propinquity

Electronic Propinquity Theory (EPT) explains how interpretive effort varies with channel features and message complexity (Korzenny 1978; Korzenny and Bauer 1981). Factors such as bandwidth, cue availability, cue alignment, and message complexity shape how difficult a message is to interpret. Media Richness Theory (Daft and Lengel 1986) offers a similar view, but EPT highlights that cues help only when they point in the same direction; misaligned cues increase effort.

In text-based communication, users rely on expressive orthography—elongation, emojis, punctuation, patterned spelling—to approximate prosody (Provine, Spencer, and Mandell 2007; Herring and Androutsopoulos 2015; McCulloch 2019). These cues can clarify affect (Rodríguez-Hidalgo, Tan, and Verlegh 2017) or serve relational functions (Hancock 2004; Walther 1992). Yet they also frequently signal sarcasm: emoticons used tongue-in-cheek (Thompson and Filik 2016), trailing punctuation (Hancock 2004), dramatic repetition, or sequences of positive emojis used ironically (Danesi 2016).

This creates a theoretical tension. *If the same cues intensify literal affect yet also index sarcastic inversion, do they help or hinder interpretation when the text is non-literal?* EPT predicts increased effort when cues contradict text; richness perspectives predict benefits from added cues. No prior controlled study has jointly manipulated sarcasm and eNVC presence. To disentangle these effects, we ask:

**RQ2a: Do eNVCs help readers accurately infer author-intended emotions in microblog posts?**

**RQ2b: Does sarcasm moderate the effectiveness of eNVCs in supporting emotional interpretation?**

### Encoding and Decoding

Encoding concerns how communicators produce signals; decoding concerns how receivers interpret them (Burgoon, Manusov, and Guerrero 2021). Hall's framework (Hall and Knapp 2013; Hall, Horgan, and Murphy 2019) emphasizes that decoding is inference-driven and shaped by expectations and norms. In face-to-face contexts, encoding involves gestures, facial movement, and prosody; online, many of these cues appear through symbolic approximations such as capitalization, repetition, expressive punctuation, stylized spelling, and emojis.

Digital prosody compresses continuous acoustic variation into discrete textual proxies ("soooo good," "WHAT??"). This flattening produces ambiguity: exclamation points may signal excitement or hostility; elongation may be affectionate, playful, mocking, or sarcastic; all-caps may indicate enthusiasm or anger. Readers diverge substantially in tonal inferences (McCulloch 2019; Krishnan and Hunt 2019).

Ambiguities shape decoding. Without cues, readers rely more on platform norms or expectations. When cues are *incongruent* with content, as in irony or mixed-valence expressions, digital prosody may increase cognitive load rather than reduce uncertainty. These cases reflect classic decoding challenges in which eNVCs add information but do not resolve affective contradiction. Sarcasm therefore represents a boundary condition: cue–text alignment is unstable, and added cues may not improve accuracy. This is consistent with theories emphasizing interpretive effort under contradictory signals. To capture these interpretive processes directly, we ask:

**RQ3: How do people interpret emotions through eNVCs in microblog posts?**

Taken together, these questions motivate our mixed-method design: (1) a nonverbal-theoretically grounded taxonomy of eNVCs; (2) an experiment testing whether eNVCs improve decoding accuracy and under which boundary conditions; and (3) focus groups analyzing how users reason through cues in context.

## Method

We conducted three studies to investigate the encoding and decoding of electronic nonverbal cues (eNVCs) in text-based social media communication. All studies were approved by the authors' university Institutional Review Board. User studies were conducted online, with participants recruited from Prolific.

Prolific was chosen as a recruitment platform because it provides access to a diverse pool of participants across OECD countries, which aligns with the global usage of X (formerly Twitter) and our interest in capturing heterogeneous audience perceptions. Prolific's international reach across OECD countries supported our goal of sampling heterogeneous audiences. We aimed for gender balance and a broad age range. Participants provided informed consent and were compensated in line with Prolific's fair payment standards. Basic demographics and platform-usage patterns were collected for descriptive reporting; no personally identifying information was retained.

### Content Analysis (Study 1)

Study 1 developed an exploratory codebook of electronic nonverbal cues (eNVCs), drawing on prior communication research and an initial content analysis of microblog posts. We used posts from *Vent*, a social media platform where users tag each post with an author-selected emotion label from a fixed set of moods (Lykousas et al. 2019). This built-in labeling makes Vent uniquely suited for validating automated cue detection against author-intended affect. The Vent corpus used here was previously released by Lykousas et al. (2019) and is publicly available. The codebook informed a Python regex library for scalable detection (Table 3).

**Dataset curation.** We curated a three-day sample from Twitter/X Trends (4–7 December 2022), empha-sizing globally salient topics likely to contain dense eNVC usage. From 4,021 tweets, we retained 118 English posts containing at least one eNVC. Because the goal of Study 1 was to develop and refine the taxonomy; this purposive sample prioritized cue diversity over volume. The resulting taxonomy was subsequently validated at scale using the Vent corpus through the regex pipeline described below.

**Annotation.** Observed cues were organized using classic nonverbal communication schemata into two text-suited domains: *kinesics* (textual depictions of bodily displays, emoji/kaomoji, stage directions) and *paralinguistics* (prosodic proxies such as capitalization, elongation, and interjections).

**Regex development.** To scale coding, we implemented a Python regex pipeline targeting punctuation, elongations, capitalization, emojis, and other textual markers. Patterns were drafted from corpus examples, refined through literature consultation and iterative prototyping, with generative assistance used only for pattern suggestions.

**Validation.** Using Vent posts, we iteratively tested regex matches and manually reviewed stratified samples. False positives (e.g., acronyms flagged as shouting) informed successive refinements until acceptable precision was reached across categories (Table 1).

## Survey Experiment (Study 2)

Study 2 tested whether eNVCs improve readers' ability to recover author-intended emotions in microblog posts, and whether this benefit holds for sarcastic posts. Using X-style stimuli with ground-truth Vent labels, we crossed eNVC presence (present vs. removed) with literality (literal vs. sarcastic) across five emotions.

**Participants** We recruited 513 English-fluent X users via Prolific, spanning Gen Z to Boomers and balanced by sex. Eligibility required English fluency and active platform use. All procedures received IRB approval.

**Materials** Stimuli were drawn from *Vent* and annotated using the Study 1 regex library. Sarcasm labels were generated in two stages. First, Llama 3.3 (70B, via Groq API) classified each candidate post as literal or sarcastic using a few-shot prompt with five annotated examples per category. Second, two researchers independently reviewed all AI-generated labels against the original post text and Vent emotion tag, resolving disagreements through discussion. This two-stage procedure ensured that sarcasm classifications reflected human judgment rather than model artifacts. A balanced 2 × 2 set (literality × cue presence) was constructed across five emotions, with 32 items shown per participant.

**Measures.** Accuracy was coded as whether participants selected the author-labeled emotion. The "Uncertain" option was analyzed as an indicator of perceived ambiguity.

**Procedure** Participants completed a randomized Qualtrics task displaying each post in X-style format. All participants viewed items from each condition. The task took about 8 minutes; participants were debriefed and compensated. Participants were compensated $1.50 for their responses.

## Focus Groups (Study 3)

Study 3 explored how users interpret and discuss eNVCs in context.

**Participants** Screening via Prolific yielded 159 eligible respondents; 25 participated across six sessions. Screening required frequent X use, English fluency, and an eNVC recognition check. The final sample was balanced by sex with a mean age of 25.

**Materials** Participants completed a brief eNVC recognition task and then discussed posts varying in cue type and ambiguity. Stimuli were sampled from the Study 1 corpus and reformatted for presentation.

**Procedure** Six 45–60 minute text-only Zoom sessions simulated microblog-style interaction. For each post, participants described the topic, inferred emotions, and identified cues. The moderator probed for disagreements, platform norms, and cultural interpretations. Sessions were transcribed. Participants were compensated $15.00 for their participation.

**Analysis** Transcripts were thematically coded to identify how participants used textual cues, negotiated ambiguity, and invoked platform norms when inferring emotion.

# Results

## Study 1

### A Taxonomy of eNVCs on Social Media

Building on prior work documenting how users adapt capitalization, elongation, punctuation, and emojis to signal affect and stance online (Krishnan and Hunt 2019; McCulloch 2019; Provine, Spencer, and Mandell 2007; Thompson and Filik 2016), we translated canonical nonverbal communication (NVC) systems into their electronic equivalents. Following (Burgoon, Manusov, and Guerrero 2022), we organize electronic nonverbal cues (eNVCs) in text-based microblogs into two domains: **kinesics** and **paralinguistics**.

- **Kinesics**: textualized gestures, touch cues, facial displays, eye movements, and emotion-conveying emoji.
- **Paralinguistics**: vocalizations, vowel and consonant elongation (prosodic lengthening), capitalization (intensity), expressive punctuation (rhythm and emphasis), and mixed-case alternation (pitch modulation).

The subcategories below reflect how traditional NVC forms manifest electronically. Tables 1 and 2 provide formal definitions and examples.

**Kinesics.** Kinesics refers to communication through bodily movement, including gestures, facial expressions, posture, and eye behavior (Birdwhistell 1952; Burgoon, Manusov, and Guerrero 2021). Classic typologies differentiate *emblems*, *illustrators*, *regulators*, *adaptors*, and *affect displays* (Ekman and Friesen 1969). In text-based CMC, these cues appear in stylized or symbolic form. Because electronic enactment requires intentional selection or typing, and because categories often overlap in practice, we collapse these into five applied eNVC subtypes:

- **Body movements (illustrators/emblems).** Textual or emoji-based actions such as 👋, *kicks*, 🧘. These operate either independently (emblems) or alongside text (illustrators).
- **Touch (haptics).** Text-based or emoji-based touch such as *hugs*, 🫂, 🤝. These functions are consistently deliberate and signal intimacy or affiliation.
- **Eye movements (oculesics).** Rendered via winks, kaomoji, or text (*squints*). We code these within kinesics due to their close functional relation to facial display.
- **Facial expressions.** Emoji and kaomoji representing facial affect (😊, 🥱, (˘˘˘)).
- **Emotion-conveying emoji.** Non-facial emoji that convey affect by context (✨, 🎉, 🎶) (Kralj Novak et al. 2015; Gesselman, Ta, and Garcia 2019).

**Paralinguistics.** Paralinguistics refers to the vocal qualities of speech, such as loudness, pitch movement, prominence, rhythm, and vocalizations, that shape affective interpretation alongside lexical content (Burgoon, Manusov, and Guerrero 2021; Hall, Horgan, and Murphy 2019). Research on prosody demonstrates that these cues are gradient, continuous, and multidimensional (Ogden 2017; Kochanski et al. 2005; Cole 2015; Ladd 2008). They play a central role in signaling emphasis, stance, and emotion (Scherer 2003).

In text-only environments, users cannot reproduce prosody acoustically. Instead, they construct *digital prosody*: stylized orthographic practices that readers interpret through cultural and genre-specific norms rather than through any phonetic equivalence (Crystal 2006; Herring and Androutsopoulos 2015; McCulloch 2019). This distinction matters: such cues evoke prosodic *functions*, not prosodic *forms*.

Guided by this perspective, we treat paralinguistic eNVCs as *perceptual* analogues of prosodic features. Our taxonomy groups these cues into three communicatively meaningful categories:

- **Vocalics (approximated vocalizations).** Interjections, laughter markers, and elongated spellings (LOL, ugh, hmmm, heyyy) evoke vocal sound or duration (Provine, Spencer, and Mandell 2007; Baron

| Kinesics sub-category | Elaboration and Example |
| --- | --- |
| Body movements | Use of emoji or text in parentheses to show bodily movement or action. Examples: 👋, 🧘, *kicks*, 💁. |
| Touch | Use of text in parentheses, or touch-related emojis, to signal physical contact. Examples: 🤝, *holds hand*, *hugs*, *leans in*. |
| Eye movements | Emojis, emoticons, or text depicting gaze or eye gestures. Examples: 😉, ;), *squints*. |
| Facial expressions | Emojis or kaomoji showing affective facial states. Examples: 😊, 🥱, (˘˘˘), XD. |
| Emotion-conveying emojis | Emojis without explicit facial features that still convey affect through symbolic or contextual use. Examples: ✨, 🎶, ❤. |

Table 1: Subcategories of Kinesics eNVCs with elaborations and examples.

2002). These signals frequently index informality, embodied reaction, or affective stance.

- **Intensity cues.** Capitalization and repeated punctuation (YES!, NO!!!!, ???) visually represent heightened emotional force, urgency, or emphasis (Herring and Androutsopoulos 2015). Their communicative value reflects shared genre conventions rather than acoustic intensity.
- **Patterned stylization (contour-like cues).** Mixed-case strings, alternating patterns, and stylized repetition (ooOoOoh, Oh GREAT, heyyyy) create visual rhythms that readers interpret as playful, emphatic, or sarcastic (McCulloch 2019; Tagliamonte 2016). These cues function as symbolic approximations of intonational contour and pragmatic prominence (Bolinger and Bolinger 1986; Ladd 2008).

We therefore conceptualize digital prosody not as a literal encoding of acoustic features, but as a set of culturally shared semiotic resources that users mobilize to convey affect, emphasis, and stance in microblog contexts. This interpretation aligns with linguistic analyses emphasizing that online prosodic cues work through inference, genre norms, and stylistic expectation (Crystal 2006; Herring and Androutsopoulos 2015; McCulloch 2019).

**Affect displays.** Affect displays, while not a distinct structural category, cut across kinesics and paralinguistics. They denote the functional use of eNVCs to express emotion (e.g., a 😍 as oculesics *and* an affect display, or NOOO!!!! as both volume and an affect display). We coded affect displays as an overlapping dimension

| Paralinguistics subcategory | Elaboration and Example |
|---|---|
| Vocalics | Written forms that evoke sounds or vocalizations. Examples: LOL, yawn, hmmm, grrr, ughhh, heyyy. |
| Volume | Use of capitalization or repeated punctuation to signal intensity or loudness. Examples: YES!, absolutely NOT, !!!, ??. |
| Pitch | Expressive elongations or alternations of letters and punctuation to indicate rising/falling intonation. Examples: ooOoOoh!, hey!!!, oh great!. |

Table 2: Subcategories of Paralinguistics eNVCs with elaborations and examples.

noted alongside the primary category.

Ambiguous or cross-cutting cases (e.g., 😍 as both oculesics and affect display) were resolved by coding to the *primary communicative function* in text, with notes for overlap. Full category definitions and examples appear in Tables 1, 2.

Appendix Table 16 summarizes collection and selection counts per topic.

**Regex Library.** The final regex library operationalizes our taxonomy into scalable, machine-readable patterns. Table 3 summarizes the categories, regex templates, and example matches. The first column demonstrates how traditional NVC categories were mapped to operational regex rules. The library is released as an open-source Python and R package at https://github.com/kokiljaidka/envc, providing a reusable toolkit for large-scale corpus analysis.

Patterns were designed to balance recall and precision: broad enough to capture the variability of microblog use (e.g., expressive elongations, multiple punctuation), yet constrained to reduce spurious matches (e.g., distinguishing acronyms from shouting).

For kinesics, we included stage-direction cues (e.g., *hug*), Unicode ranges for facial and body-part emojis, and a subset of affective symbols (e.g., ❤️, ✨). For paralinguistics, we targeted vocalics lexicons (*lol*, *yawn*), capitalization spans, repeated punctuation, elongations, and alternating case forms that mimic prosodic shifts. As an illustration, the post *"I have so much homework & I'm overthinking.. 😔 gawd I'm such a mess..."* (author-labeled emotion: stressed) was matched by the regex library for repeated punctuation ("..."), emoji (😔), and vocalics ("gawd"), demonstrating how multiple eNVC categories co-occur within a single post. Additional matched examples appear in Appendix B.

| Domain | Subcategory | Regex Pattern | Example Matches |
|---|---|---|---|
| Kinesics | Body / Touch | (hug\|wave\|frown\|smile\|clap) | *hug*, *frowns* |
|  | Facial / Eye | Unicode ranges for faces: [0001F600-0001F64F] | 😀, 😂 |
|  | Body-part emoji | Unicode ranges for body: [0001F400-0001F4FF] | 💪, 👏 |
|  | Emotion-conveying emoji | Heart/symbol ranges: [0001F490-0001F9E1] | ❤️, ✨, 💐 |
| Paralinguistics | Vocalics | (lol\|yawn\|ugh+\|hmmm+) | lol, yawn, ughhh |
|  | Volume | [A-Z]2, | THIS, STOP |
|  | Volume (punct.) | !!+ or +| | !!!, ??? |
|  | Pitch (elongation) | ()2, | soooo, noooo |
|  | Pitch (alt. case) | ([A-Z][a-z])2, | HiYa, LoL |

Table 3: Regex patterns for detecting eNVC categories in text.

**Category frequencies.** To characterize the distribution of eNVC types at scale, we applied the regex library to the Vent sample of posts containing at least one eNVC. Paralinguistic cues were substantially more common than kinesic cues: intensity markers (capitalization, repeated punctuation) were the most frequent category, followed by vocalics, emotion-conveying emoji, and facial-expression emoji. Patterned stylization was the rarest category. These distributions inform the relative salience of cue types in microblog communication but should not be treated as population estimates given the platform-specific corpus.

## Study 2 Results

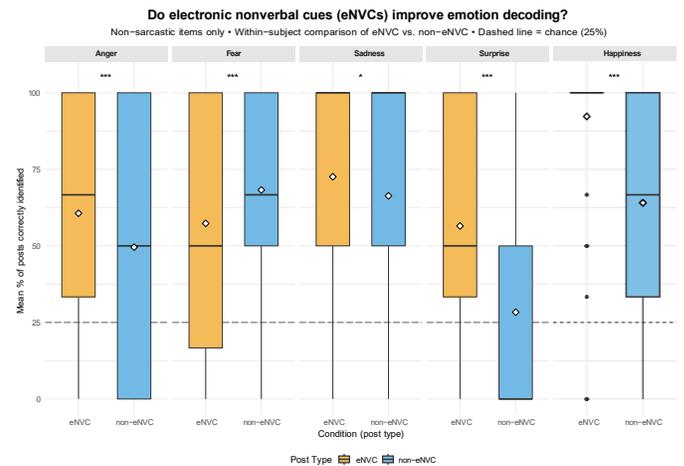

Figure 1: Within-subject comparison of decoding accuracy for microblog posts with vs. without eNVCs (in the no-sarcasm conditions) across five emotions ($N$ = 513). Dashed line indicates chance level (25%).

We began by comparing within-participant performance across the four experimental cells. Because each participant evaluated an equal number of posts in every condition, we conducted paired-samples $t$-tests on participant-level accuracy scores.

**Paired comparisons.** Participants were substantially more accurate on *non-sarcastic* posts when eN-

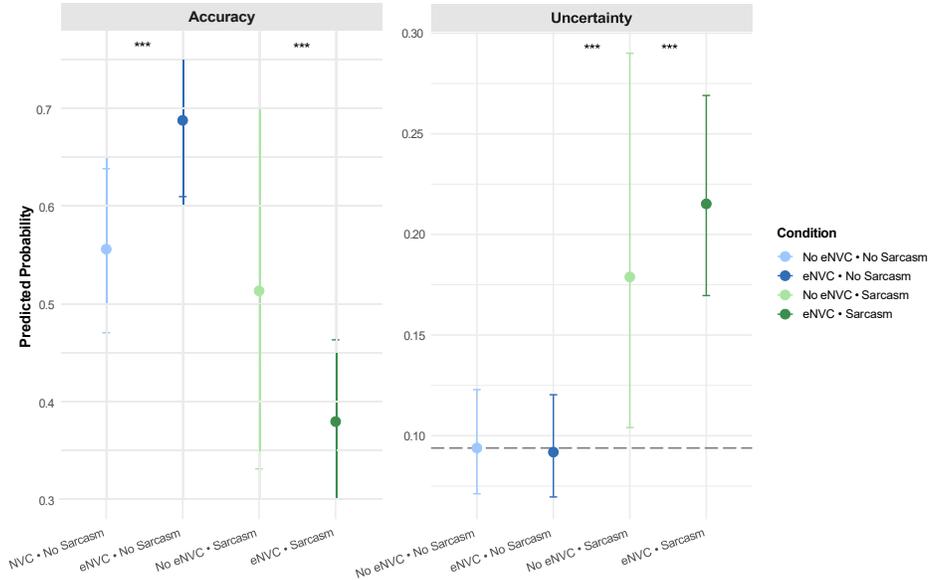

Figure 2: Mixed-effects predictions for accuracy and uncertainty.

VCs were present than when they were removed. On average and across emotions, participants answered 1.32 more items correctly (out of 8 per condition) when eNVCs were present than when they were absent ($t_{512}$ = −17.18, $p < .001$, 95% CI [−1.47, −1.17]) (see Figure 1). A similar pattern held for *sarcastic* posts: participants answered 0.92 more sarcastic items correctly when eNVCs were preserved than when they were stripped ($t_{512}$ = −12.17, $p < .001$, 95% CI [−1.07, −0.77]). These within-subject comparisons confirm that, at the participant level, eNVCs improve decoding performance for both literal and sarcastic posts, though the benefit is smaller for sarcasm.

**Mixed-effects accuracy models.** To adjust for item-level heterogeneity and fully exploit the trial-level design, we fit mixed-effects logistic regressions with random intercepts for participants and emotions, using the *non-eNVC, non-sarcastic* condition as the baseline.

The main-effects model revealed two robust findings (Table 11). First, preserving eNVCs improved accuracy: posts with eNVCs were significantly more likely to be decoded correctly than those without them ($\beta$ = 0.56, $SE$ = 0.05, $p < .001$, OR = 1.75). Second, sarcasm substantially impaired decoding ($\beta$ = −1.27, $SE$ = 0.05, $p < .001$, OR = 0.28). Demographic covariates were small and inconsistent, with the exception of a modest accuracy advantage for White participants.

A four-condition specification decomposed these effects across the full 2 × 2 structure. Figure 1a shows that, relative to the *non-eNVC, non-sarcastic* baseline:

- **Non-sarcastic + eNVC** posts were substantially more accurate ($\beta$ = 0.56, $SE$ = 0.05, $p < .001$, OR = 1.76).
- **Sarcastic + no eNVC** posts did not differ reliably from baseline ($\beta$ = −0.17, $SE$ = 0.42, $p$ = .69), indicating increased variability in sarcastic decoding.
- **Sarcastic + eNVC** posts were markedly less accurate than baseline ($\beta$ = −0.71, $SE$ = 0.05, $p < .001$, OR = 0.49).

Estimated marginal means show highest accuracy for non-sarcastic eNVC posts and the lowest accuracy for sarcastic posts regardless of cue presence. These results demonstrate that eNVCs improve decoding in literal contexts but cannot fully compensate for the semantic inversion introduced by sarcasm.

**Perceived ambiguity (uncertainty).** We next analyzed selections of the "Uncertain" response option, which reflects subjective ambiguity rather than categorical accuracy. In the main-effects model, eNVCs did not reliably reduce uncertainty ($\beta$ = −0.01, $SE$ = 0.07, $p$ = .83), whereas sarcasm strongly increased uncertainty ($\beta$ = 0.99, $SE$ = 0.06, $p < .001$, OR = 2.69). Thus, sarcastic posts were nearly three times more likely to elicit explicit uncertainty than literal ones.

The four-condition uncertainty model sharpened this pattern. Figure 1b shows that, relative to the *non-eNVC, non-sarcastic* baseline:

- **Non-sarcastic + eNVC** posts showed no reliable difference in uncertainty ($\beta$ = −0.02, $SE$ = 0.07, $p$ = .73).

- **Sarcastic + no eNVC** posts elicited significantly more uncertainty ($\beta = 0.74$, $SE = 0.35$, $p = .035$, $OR = 2.10$).
- **Sarcastic + eNVC** posts showed the largest increase in uncertainty ($\beta = 0.97$, $SE = 0.06$, $p < .001$, $OR = 2.65$).

Thus, although eNVCs strengthen accuracy for literal items, they do not diminish the subjective ambiguity introduced by sarcasm.

In summary, across analytic approaches, eNVCs consistently improved accuracy for literal posts and produced modest gains even for sarcastic ones in paired comparisons. However, sarcasm sharply reduced categorical accuracy and substantially heightened subjective ambiguity. Under sarcasm, eNVCs failed to offset these effects—they neither lowered uncertainty nor restored correct interpretation—illustrating a clear boundary condition on the communicative value of textual nonverbal cues.

**Interpretation of findings** Study 2 examined whether eNVCs improve readers' ability to decode author-intended emotions in microblog posts. The within-subject design yielded a strong and consistent advantage for eNVCs. A paired-samples analysis showed that participants identified emotions more accurately when posts contained eNVCs than when the same posts were presented without them. Across the 30 items, accuracy was significantly higher in the eNVC condition. These effects held even though participants who performed well in one condition tended to perform well in the other, indicating that eNVCs improve performance beyond baseline decoding skill. Emotion-specific analyses showed a broadly consistent eNVC advantage, though the magnitude varied. Improvements were largest for happiness, surprise, and fear - emotions that in text-only form tend to rely on contextual or stylistic disambiguation. Gains were smaller for sadness and anger, which are more reliably signaled through lexical content alone. This pattern aligns with prior work showing that paralinguistic and prosodic markers carry more diagnostic value in positive or mixed-valence expressions than in strongly negative disclosures (Coyle and Carmichael 2019).

Mixed-effects models that accounted for participant- and emotion-level heterogeneity corroborated these results. Relative to the reference condition (non-eNVC, non-sarcastic), the odds of correct decoding were substantially higher for non-sarcastic items containing eNVCs ($OR = 1.76$), and substantially lower for sarcastic items regardless of cue presence ($OR = 0.49$ for eNVC sarcasm; Table 12). These estimates reflect a robust main effect of eNVCs and a large negative effect of sarcasm.

Estimated marginal means provide further clarity. For literal, non-sarcastic posts, eNVCs raised the probability of correct decoding. For sarcastic posts, however, accuracy remained low overall, reflecting the inherent difficulty of resolving deliberate text–affect misalignment.

Analyses of perceived ambiguity ("uncertain" responses) revealed a complementary pattern. For non-sarcastic items, uncertainty rates were uniformly low and unaffected by eNVC presence. For sarcastic items, uncertainty was substantially higher overall, and highest for sarcastic items with eNVCs—indicating that additional cues do not resolve, and may even heighten, the ambiguity introduced by irony.

Taken together, these results support a two-part conclusion. First, eNVCs reliably enhance emotion recognition for literal or straightforward expressions in microblog posts. Second, their benefits attenuate sharply in sarcastic or affectively incongruent contexts: while readers may feel less uncertain in some paired-sample comparisons, eNVCs do not improve categorical accuracy once item-level variability is modeled. These boundary conditions highlight contexts in which digital nonverbal cues facilitate interpretation and those in which they cannot overcome deeper semantic contradictions. We have explored these issues qualitatively in Study 3.

## Study 3

Twelve posts were analyzed across six text-based focus group sessions, with each post discussed in at least two sessions (Appendix H). Transcripts were qualitatively coded at multiple levels: individual answers, conversational threads, and group consensus. Four interpretive patterns emerged, each illuminating a distinct mechanism through which readers decode (or fail to decode) affect from textual cues.

**Cue convergence disambiguates affect.** When multiple eNVCs aligned in valence, participants reported high interpretive confidence, consistent with Study 2's accuracy gains for literal eNVC posts. Critically, readers did not simply register individual cues; they treated co-occurring cues as mutually reinforcing. For *"Just finished reading Fahrenheit 451... DAMN... I really enjoyed reading it 👌"*, one participant explained, "the word 'DAMN' plus the emoji showcases that they are impressed and mind blown," and another added, "the emoji seals it, it makes clear they're positive, not annoyed." This convergence effect contradicts Walther and Bazarova's (Walther and Bazarova 2008) prediction that cue complexity reduces clarity; instead, redundant cues narrowed the interpretive space. Participants also parsed *distinct affective layers* from converging cues. For *"EXAAAAAAAAAMS 😢"*, one noted, "the emoji shows sadness, but the stretched letters feel like frustration," and another added, "it's also like panic, like they're overwhelmed, not just sad." Here, kinesic and paralinguistic cues jointly specified a blended emotional state (frustration + anxiety) that neither cue alone would have conveyed, suggesting that eNVCs do not merely amplify a single emotion but can scaffold *affective differentiation*.

Similarly, for *"Today is Friday!!!!!!! thannkkkk gooood"*, a participant reasoned: "the exclamation marks and the repeated use of the same letters denotes excitement…while you read it you can hear a tired person being thankful for rest." The metaphor of *hearing* a voice through orthography illustrates how readers recruit prosodic imagery to resolve textual cues: a process akin to intertextual reading (Orr 2003), where meaning is constructed not from any single marker but from the gestalt of overlapping signals.

**Cue excess triggers sarcasm inference.** The same intensity markers that clarified literal affect became ambiguous or ironic when their density exceeded genre expectations. This pattern provides qualitative grounding for Study 2's finding that eNVCs under sarcasm *reduced* accuracy rather than merely failing to help. In *"I think my record player just came in the mail 😍😍😍😍 THANK YOU FEDEX"*, one participant read the repeated emojis as genuine intensification ("the extra emojis mean the emotion is really strong"), while another countered, "I'd use that many emojis mostly if I were being sarcastic lol." For *"Oh I am soo tiiiired ughh wow..."*, participants converged on annoyance: "the repeats show they're sarcastic, mocking how people always say they're tired." These exchanges reveal a threshold effect: moderate eNVC use signals sincerity, but excessive use activates an irony frame (Hancock 2004; Danesi 2016). Importantly, participants *disagreed* about where the threshold lay, which helps explain why sarcastic eNVC posts in Study 2 produced both low accuracy and high uncertainty: readers apply different genre calibrations to the same stimulus.

**Absence as a cue: flat styling signals affective weight.** A theoretically novel finding was that participants inferred emotion not only from what was present but from what was *missing*. This goes beyond standard accounts of cue richness, which focus on what channels provide; here, the interpretive resource is an expectation violation. In *"When the guy you like starts talking to another girl.. 👌 "*, one participant reasoned:

> "Unless they express anger in a calm way, then wow. I think anger would be expressed with capital letters, angry emojis…all in all I think the text would [have been] dramatised."

The absence of capitalization or exclamation marks ruled out anger, leading the group toward disappointment or resigned sadness. Similarly, for *"I'm shit at everything I do 👍"*, a participant observed: "there's no exaggerated punctuation, no exclamation marks no nothing, it's so matter of fact." Flat styling was read as emotional weight: too heavy for performative complaint, too plain for humor. This *diagnostic absence* implies that readers maintain implicit expectations about what "normal" emotional expression looks like on a platform, and interpret deviations *in either direction*, i.e., excessive cues or conspicuously missing cues, as meaningful. For EPT, this suggests that interpretive effort is shaped not just by what the channel transmits but by the gap between what is transmitted and what is expected.

**Negativity bias under ambiguity.** When cue–text alignment was unclear, participants consistently defaulted to more negative interpretations than the author labels suggested. For the "another girl" post, readings ranged from disappointment and jealousy to anger and sarcasm, with ellipses flagged as ironic: "It's the two dots before the emoji that show sarcasm." Self-deprecating posts drew particular suspicion. Regarding *"I'm shit at everything I do 👍"*, one participant inferred strategic intent: "The person is masking a negative self emotion as a sarcastic tweet…trying to fish compliments to lift them up." For *"uhhh... all my friends hate me... haha kill me"*, others speculated the author was "bragging a bit while supposedly showing some bad traits." These responses reveal a systematic interpretive asymmetry: under uncertainty, readers attribute darker motives (manipulation, insincerity, concealed distress) rather than taking posts at face value. This negativity bias echoes findings on cynical interpretation in CMC (Ferguson 2021; McCreery and Krach 2018) and has practical implications for content moderation systems that rely on surface-level cue detection without modeling the audience's tendency to "read down."

**Summary.** Across these four patterns, Study 3 reveals that eNVC decoding is not a simple feature-recognition task but an inferential process shaped by cue convergence, genre-calibrated thresholds for sincerity, expectations about absent cues, and a systematic negativity bias under ambiguity. These mechanisms help explain both the strong accuracy gains for literal posts in Study 2 (convergence narrows interpretation) and the sharp accuracy losses under sarcasm (excess triggers competing frames). Together, they extend encoding–decoding models by showing that the interpretive value of eNVCs depends not on cue presence alone but on the fit between cues, content, and platform-specific expectations about how emotion *should* be expressed.

## Discussion

Across three studies, this work shows how electronic nonverbal cues (eNVCs) function in microblog communication and how audiences interpret them. We develop a taxonomy grounded in nonverbal communication theory, demonstrate experimentally that eNVCs generally improve emotion recognition, and document how users reason with these cues, including their responses to the absence of expected cues. The findings refine accounts of cue richness, interpretive effort, and social presence by identifying when additional cues facilitate meaning and when benefits diminish, particularly under sarcasm or mixed affect. They also broaden eNVC research beyond emojis to include symbolic, structural, and paralinguistic markers.

Methodologically, we integrate content analysis, automated text processing, experimentation, and qualitative inquiry. Study 1 introduced a scalable regex pipeline; Study 2 provided cross-national causal evidence; and Study 3 illustrated how readers integrate cues with platform norms and expectations. Together, these approaches show the value of mixed methods for understanding digital nonverbal behavior.

**From Translation to Validation** A central contribution is translating foundational nonverbal communication categories into electronic form. Study 1 introduced a taxonomy distinguishing *kinesics* (textual gestures, emojis) and *paralinguistics* (vocalics, intensity, elongation). Operationalizing this taxonomy via regular expressions enabled scalable detection for affective computing and moderation tasks.

Studies 2 and 3 validate the taxonomy. In Study 2, participants more accurately identified author-intended emotions when original eNVCs were present. Effects were substantial for non-sarcastic items, consistent with paired *t*-tests and marginal means. For sarcastic items, eNVCs reduced accuracy and increased perceived ambiguity. Study 3 further showed that users integrate multiple cues, debate ambiguous cases, and attend to flat or minimal styling as meaningful. These convergent findings confirm that the taxonomy captures expressive resources users actively draw on when decoding affect.

**Cue Richness, Interpretive Effort, and Boundary Conditions** The gradient nature of spoken prosody allows listeners to calibrate affective intensity continuously; textual prosody, by contrast, compresses this into discrete symbols. This compression has a theoretical consequence: eNVCs can *narrow* the space of plausible interpretations without *eliminating* ambiguity. Our results specify when narrowing succeeds and when it fails.

For literal posts, eNVCs function as predicted by media richness accounts (Daft and Lengel 1986): added cues reduce interpretive candidates and raise accuracy (OR = 1.76). The mechanism suggested by Study 3 is *cue convergence*, when multiple eNVCs (e.g., elongation + emoji + punctuation) point toward the same affect, readers treat the conjunction as confirmatory, not redundant. This contradicts Walther and Bazarova's (Walther and Bazarova 2008) prediction that cue complexity reduces clarity and instead supports an additive model of digital expressiveness.

For sarcastic posts, however, the same cues that clarify literal affect become sources of interference. Accuracy dropped below the non-eNVC baseline (OR = 0.49), and perceived ambiguity more than doubled. This is not merely a failure of cues to help: it is an active cost. We interpret this through EPT (Korzenny 1978; Korzenny and Bauer 1981): when cues and text point in opposing affective directions, each additional cue adds a competing signal that the reader must adjudicate, raising interpretive effort rather than reducing it. The focus groups corroborated this: participants described elongation and emoji repetition in sarcastic posts as confusing precisely because the same markers were reliable in literal contexts.

This pattern, i.e., facilitation under alignment, interference under misalignment, constitutes a *coherence-dependent* model of cue utility. It refines both media richness and EPT by specifying that cue *quantity* is subordinate to cue–text *coherence* as a predictor of interpretive success.

**Implications for Design and Computation** These findings carry practical implications for three domains. First, for *affective computing*, sentiment and emotion classifiers that treat eNVCs as uniformly positive signals will systematically misclassify sarcastic content; models should incorporate cue–text coherence features. Second, for *content moderation*, the negativity bias documented in Study 3, where ambiguous cues default to negative readings, suggests that flagging systems tuned to surface-level cue intensity may over-flag ironic or playful posts. Third, for *interface design*, platforms could support affective clarity by offering structured tone indicators (e.g., tone tags such as /s or /gen) that reduce reliance on ambiguous paralinguistic cues. These recommendations are preliminary and require validation in deployed systems.

**Decoding Beyond Presence: Absence, Thresholds, and Negativity Bias** Study 3 extends the coherence-dependent model by revealing three interpretive mechanisms that operate alongside cue convergence. First, participants treated the *absence* of expected cues as diagnostically meaningful: flat styling signaled emotional weight or resignation, not neutrality. This implies that readers maintain implicit baselines for "normal" eNVC density on a platform and interpret deviations in either direction (either excessive or conspicuously absent cues) as affectively loaded. For EPT, this means interpretive effort is shaped not only by what the channel transmits but by the gap between transmission and expectation. Second, focus group disagreements about whether cue repetition was sincere or ironic revealed a *threshold effect*: moderate eNVC use reads as genuine, but excess activates a sarcasm frame, with the threshold varying across readers. This inter-individual variation helps explain the high uncertainty in Study 2's sarcastic conditions, as participants applied different genre calibrations to identical stimuli. Third, under ambiguity, readers defaulted to negative interpretations, attributing manipulation, insincerity, or concealed distress rather than taking posts at face value (Ferguson 2021; McCreery and Krach 2018). This negativity bias has direct consequences for affective computing and moderation systems that rely on surface-level cue detection without modeling audience interpretive tendencies.

Taken together, these mechanisms show that eNVC decoding is an inferential process governed by cue-text coherence, genre-calibrated expectations, and systematic interpretive asymmetries, and not a simple feature-recognition task. The project contributes theoretically by specifying when and why cues help or complicate interpretation, and practically by offering a validated, scalable toolkit for detecting eNVCs.

**Limitations** Several limitations warrant discussion. First, all stimuli are English-language microblogs. eNVC conventions are likely language- and script-dependent: elongation patterns differ in logographic writing systems, emoji semantics vary cross-culturally, and punctuation norms are not universal. Although our international participant sample captured some cross-national variation, cultural differences in cue interpretation were not experimentally manipulated. The threshold and negativity-bias effects documented in Study 3 may operate differently in high-context communication cultures where indirectness carries different pragmatic weight. Second, author-labeled emotions from Vent serve as ground truth, but self-reports may not always match the affect perceived by third-party readers; this gap is inherent to any encoding–decoding study and is partially addressed by the focus groups, which revealed how third-party readings diverge from author intent. Third, the taxonomy focuses on orthographic and symbolic cues in text-only posts. Social media increasingly supports multimodal expression through GIFs, short video, voice messages, memes, and hashtag pragmatics, all of which carry nonverbal meaning. The cue-convergence mechanism identified in Study 3 likely operates across modalities (e.g., a laughing GIF paired with elongated text), but testing this requires different detection methods. Fourth, the Study 1 taxonomy was developed from a purposive sample of 118 posts; while the regex pipeline was validated on the larger Vent corpus, the initial category set may not capture all eNVC forms used across platforms and communities. Fifth, Study 3's finding that cue excess triggers sarcasm inference suggests that the sincerity–irony threshold varies across individuals and likely across platform norms; our six focus groups cannot establish where this threshold falls for broader populations. Despite these constraints, the convergence of experimental and qualitative evidence indicates that the coherence-dependent model and its boundary conditions are robust. Future work should test the taxonomy in multilingual corpora, extend it to multimodal cues, and assess whether the threshold and negativity-bias effects replicate across linguistic, cultural, and platform contexts.

## Conclusion

This project examined how eNVCs are encoded and decoded through three complementary studies. Study 1 introduced a theoretically grounded taxonomy and scalable regex pipeline. Study 2 offered experimental evidence that eNVCs improve recognition of author-intended emotions in literal posts, with effects substantially attenuated by sarcasm. Study 3 identified the interpretive mechanisms underlying these patterns: cue convergence narrows interpretation for literal posts, cue excess activates competing sarcasm frames, diagnostic absence carries affective weight, and ambiguity triggers a systematic negativity bias.

The central theoretical contribution is a *coherence-dependent* model of cue utility. Across studies, cue-text alignment predicted interpretive success. When eNVCs converged with lexical content, accuracy rose substantially (OR = 1.76); when they accompanied sarcastic or contradictory text, accuracy fell below baseline (OR = 0.49) and perceived ambiguity more than doubled. This pattern refines both media richness and Electronic Propinquity Theory by specifying that added cues assist interpretation only when they reinforce, rather than compete with, textual affect.

Methodologically, the studies demonstrate the value of combining computational, experimental, and qualitative approaches. The taxonomy enables scalable detection; the experiment isolates causal effects; and the focus groups reveal mechanisms: threshold effects, absence-based inference, negativity bias, that are invisible in aggregate accuracy data.

In sum, this work contributes a theoretically grounded taxonomy and detection toolkit for eNVCs, controlled evidence of their interpretive value and limits, and qualitative insight into how users reason through digital prosody, including when cues clarify, when they complicate, and when their absence speaks loudest.

**Acknowledgment:** This work was supported by the Singapore Ministry of Education AcRF TIER 3 Grant (MOET32022-0001), Tier 1 programme (WBS A-8000231-01-00) and A*STAR OTS A-8003288-00-00. We're grateful to feedback from the Journal Club at CNM, NUS.

# Paper Checklist

Please note that we compiled the paper using LuaLaTeX because several results rely on emoji-based stimuli. As a result, some minor formatting differences (e.g., font rendering) may appear relative to the standard pdfLaTeX output. We have ensured that all content, figures, and layouts follow the submission guidelines as closely as possible. We appreciate your understanding.

1. For most authors...

   (a) Would answering this research question advance science without violating social contracts, such as violating privacy norms, perpetuating unfair profiling, exacerbating the socio-economic divide, or implying disrespect to societies or cultures? Yes. The study uses public social-media datasets and consented online participants, does not enable individual targeting, and aims to improve understanding of emotion decoding in text-based communication.

   (b) Do your main claims in the abstract and introduction accurately reflect the paper's contributions and scope? Yes. The abstract and introduction clearly state the three studies, the eNVC taxonomy and regex toolkit, the survey experiment, and the focus-group analysis, and these correspond to the reported results.

   (c) Do you clarify how the proposed methodological approach is appropriate for the claims made? Yes. The paper motivates the mixed-method design—content analysis with a regex pipeline, a within-subjects survey experiment, and qualitative focus groups—as appropriate to answer the three research questions.

   (d) Do you clarify what are possible artifacts in the data used, given population-specific distributions? Yes. The paper discusses that the data are English-language, primarily Western/Anglophone, and platform-specific, and notes that interpretations of eNVCs may vary across linguistic and cultural communities.

   (e) Did you describe the limitations of your work? Yes. The limitations include language and cultural scope, platform coverage, stylistic forms not modeled by the taxonomy, and the boundary conditions introduced by sarcasm and mixed affect.

   (f) Did you discuss any potential negative societal impacts of your work? Yes. The ethical considerations section notes risks of over-generalizing eNVC interpretations, privileging Western-centric norms, and essentializing differences if cue interpretations are treated as universal.

   (g) Did you discuss any potential misuse of your work? Yes. The paper cautions against using eNVC measures as universal or diagnostic signals and emphasizes that interpretations are context- and culture-dependent, which would make misuse for stereotyping or individual profiling inappropriate.

   (h) Did you describe steps taken to prevent or mitigate potential negative outcomes of the research, such as data and model documentation, data anonymization, responsible release, access control, and the reproducibility of findings? Yes. The paper collected no personally identifiable information during its user studies. We provide detailed documentation of the regex toolkit, sampling, and analysis pipelines to support responsible reuse.

   (i) Have you read the ethics review guidelines and ensured that your paper conforms to them? Yes.

2. Additionally, if your study involves hypotheses testing...

   (a) Did you clearly state the assumptions underlying all theoretical results? Not applicable. The paper reports empirical results addressing research questions but does not present formal theoretical results.

   (b) Have you provided justifications for all theoretical results? Not applicable.

   (c) Did you discuss competing hypotheses or theories that might challenge or complement your theoretical results? Not applicable.

   (d) Have you considered alternative mechanisms or explanations that might account for the same outcomes observed in your study? Not applicable.

   (e) Did you address potential biases or limitations in your theoretical framework? Not applicable.

   (f) Have you related your theoretical results to the existing literature in social science? Not applicable.

   (g) Did you discuss the implications of your theoretical results for policy, practice, or further research in the social science domain? Not applicable.

3. Additionally, if you are including theoretical proofs...

   (a) Did you state the full set of assumptions of all theoretical results? Not applicable.

   (b) Did you include complete proofs of all theoretical results? Not applicable.

4. Additionally, if you ran machine learning experiments...
   (a) Did you include the code, data, and instructions needed to reproduce the main experimental results (either in the supplemental material or as a URL)? Not applicable. The paper does not train or benchmark predictive machine-learning models; analyses use content coding, regular expressions, and statistical modeling.
   (b) Did you specify all the training details (e.g., data splits, hyperparameters, how they were chosen)? Not applicable. No ML training is performed.
   (c) Did you report error bars (e.g., with respect to the random seed after running experiments multiple times)? Not applicable.
   (d) Did you include the total amount of compute and the type of resources used (e.g., type of GPUs, internal cluster, or cloud provider)? Not applicable.
   (e) Do you justify how the proposed evaluation is sufficient and appropriate to the claims made? Not applicable. Evaluation relies on standard statistical analyses rather than ML benchmarking.
   (f) Do you discuss what is "the cost" of misclassification and fault (in)tolerance? Not applicable.

5. Additionally, if you are using existing assets (e.g., code, data, models) or curating/releasing new assets, **without compromising anonymity**...
   (a) If your work uses existing assets, did you cite the creators? Yes. Prior work and sources for the Vent corpus, Twitter/X data, and large language models used in stimulus curation are cited.
   (b) Did you mention the license of the assets? Yes. The paper notes that social-media data are used under platform terms of service and that externally released datasets are used under their stated access conditions.
   (c) Did you include any new assets in the supplemental material or as a URL? Yes. The paper shares the eNVC taxonomy, regex patterns, and study materials (stimuli examples, moderator guides, and codebook details) to support reuse.
   (d) Did you discuss whether and how consent was obtained from people whose data you're using/curating? Yes. For Studies 2 and 3, participants were recruited via Prolific, provided informed consent, and were compensated; for public social-media data, the paper relies on existing dataset releases and platform terms.
   (e) Did you discuss whether the data you are using/curating contains personally identifiable information or offensive content? Yes. The paper notes that no personally identifying information was retained and that some stimuli include negative or self-deprecating content typical of social-media posts.
   (f) If you are curating or releasing new datasets, did you discuss how you intend to make your datasets FAIR? Not applicable. The paper does not release a new standalone dataset; it provides tools and materials built on existing corpora.
   (g) If you are curating or releasing new datasets, did you create a Datasheet for the Dataset? Not applicable.

6. Additionally, if you used crowdsourcing or conducted research with human subjects, **without compromising anonymity**...
   (a) Did you include the full text of instructions given to participants and screenshots? Yes. Sample stimuli, task instructions, and the focus-group moderator guide are provided in the appendices and supplemental material.
   (b) Did you describe any potential participant risks, with mentions of Institutional Review Board (IRB) approvals? Yes. The paper reports IRB approval, informed consent procedures, and describes the studies as minimal-risk online tasks involving exposure to everyday social-media posts.
   (c) Did you include the estimated hourly wage paid to participants and the total amount spent on participant compensation? Yes, it is reported in the Method section.
   (d) Did you discuss how data is stored, shared, and deidentified? Yes. The paper notes that no personally identifying information was retained, that data are stored in deidentified form, and that only public social-media content and anonymized transcripts are used in analysis.

**Ethical considerations.** Emotion decoding norms are culturally contingent; emojis, elongations, and textual gestures vary across linguistic and regional contexts. To mitigate this, Studies 2 and 3 recruited participants internationally while fixing the survey language to English. This captured cross-national variation while maintaining shared linguistic grounding. Our approach aligns with "difference-aware" perspectives (Wang et al. 2025), which warn against assuming uniform interpretation. Establishing a cross-cultural eNVC taxonomy requires multilingual corpora, localized focus groups, and openness to interpretive diversity to avoid essentializing communicative practices.

# Appendix A: Demographics of Study 2 Participants

A total of 513 participants completed the Study 2 task. All demographic summaries reported below reflect this final analytical sample.

## Sex and Age

The sample was nearly evenly split by sex, with 259 females (50.1%) and 254 males (49.1%). Four participants did not provide valid sex information.

Participants ranged in age from 18 to 75. Using standard generational boundaries, 25% were members of Generation Z (ages 18–26), 52% were Millennials (27–42), 16% were Generation X (43–58), and 6% were Baby Boomers (59+). These distributions indicate that although younger adults were well represented, the sample included substantial mid-life and older participants, consistent with our broad recruitment targets.

| Sex | Count (%) |
|---|---|
| Female | 259 (50.1%) |
| Male | 254 (49.1%) |
| Total (valid) | 513 |

Table 4: Sex of participants in Study 2

### Country of Residence

Participants represented more than 25 countries across six continents. The largest groups were from Spain (189), the United Kingdom (96), Italy (28), Canada (26), Poland (18), and Portugal (18). Additional participants resided in countries across Europe, Asia, Africa, North America, South America, and Oceania. Table 5 presents a full breakdown.

### Ethnicity

Participants reported their ethnicity using Prolific's simplified categories. The largest group identified as White (241, 47%). This was followed by Black (190, 37%), Asian (41, 8%), Mixed (27, 5%), and Other (12, 2%). One participant selected "Prefer not to say."

## Appendix B: Survey Experiment Items and Selected Results (Study 2)

This appendix presents example items and selected response distributions from the 30-item emotion–identification task used in Study 2. Each item consisted of a microblog post shown either with or without electronic nonverbal cues (eNVCs), and participants selected the emotion they believed the author intended to express.

### Example 1: Happiness (non-eNVC)

*Tweet:* "singing is a wonderful remedy."

*Response options:* Stressed, **Calm (correct)**, Surprised, Sad, Unsure, No emotion.

### Example 2: Happiness (eNVC)

*Tweet:* "My cat cuddled with me and I feel better. I love caaaats 😻😻"

*Response options:* Amused, **Happy (correct)**, Empty, Ashamed, Unsure, No emotion.

## Appendix C: Results: Accuracy Analyses

Table 9 reports the paired-samples t-tests comparing accuracy across eNVC conditions.

| Country | Count |
|---|---|
| Australia | 2 |
| Brazil | 6 |
| Canada | 26 |
| Chile | 8 |
| Croatia | 2 |
| Czech Republic | 1 |
| Egypt | 2 |
| Estonia | 2 |
| France | 5 |
| Germany | 6 |
| Greece | 8 |
| Hungary | 6 |
| India | 9 |
| Indonesia | 3 |
| Ireland | 4 |
| Italy | 28 |
| Kenya | 7 |
| Korea | 1 |
| Malaysia | 2 |
| Mexico | 4 |
| Morocco | 1 |
| Netherlands | 9 |
| Norway | 2 |
| Poland | 18 |
| Portugal | 18 |
| Slovenia | 2 |
| South Africa | 189 |
| Spain | 13 |
| Sweden | 2 |
| United Kingdom | 96 |
| United States | 29 |
| Vietnam | 1 |
| Not Available | 1 |
| Total | 513 |

Table 5: Country of residence of Study 2 participants

To account for item-level heterogeneity and to fully exploit the within-subject design of Study 2, we estimated generalized linear mixed-effects models predicting whether each trial was correctly decoded. All models include random intercepts for participants and emotion categories.

### Model 1: Main Effects of eNVCs and Sarcasm

Model 1 (Table 10) estimated the independent effects of preserving nonverbal cues (eNVCs) and sarcasm, adjusting for demographic covariates. The presence of eNVCs significantly increased the odds of correct interpretation ($\hat{\beta}$ = 0.559, $SE$ = 0.047, $p$ < .001, OR = 1.75). Sarcasm had a large negative effect on accuracy ($\hat{\beta}$ = −1.274, $SE$ = 0.048, $p$ < .001, OR = 0.28). Demographic covariates were small and inconsistent, aside from a modest positive effect for White participants. Random intercept variances indicate meaningful

| Ethnicity | Count (%) |
|---|---|
| White | 241 (47%) |
| Black | 190 (37%) |
| Asian | 41 (8%) |
| Mixed | 27 (5%) |
| Other | 12 (2%) |
| Prefer not to say | 1 (0.2%) |
| Total | 512 |

Table 6: Ethnicity of participants in Study 2

| Response Option | Count |
|---|---|
| Calm (correct) | 355 |
| Sad | 9 |
| Stressed | 15 |
| Surprised | 14 |
| No emotion | 35 |
| Unsure / Other | 61 |
| Total | 489 |

Table 7: Response distribution for Example 1

heterogeneity across participants and emotions.

## Model 2: Four-Condition Specification

Model 2 (Table 11) decomposed the 2 × 2 structure into four mutually exclusive conditions, using the *non-eNVC, non-sarcastic* cell as the reference category. Estimated marginal means are reported in Table 12.

Relative to this baseline:

- **eNVC, non-sarcastic** posts were substantially more likely to be decoded correctly ($\hat{\beta}$ = 0.565, SE = 0.047, $p$ < .001, OR = 1.76).

- **Non-eNVC, sarcastic** posts did not differ reliably from baseline ($\hat{\beta}$ = −0.171, SE = 0.422, $p$ = .685), indicating large variability in sarcastic decoding.

- **eNVC, sarcastic** posts were significantly less likely to be decoded correctly ($\hat{\beta}$ = −0.715, SE = 0.046, $p$ < .001, OR = 0.49), confirming the strong accuracy cost of sarcasm even when cues are preserved.

Consistent with Model 1, eNVCs improve accuracy for literal posts but cannot offset the inherent semantic inversion of sarcasm.

# Appendix D: Results: Uncertainty Analyses

In addition to categorical accuracy, we examined participants' subjective uncertainty. For each trial, choosing the "Uncertain" option indicates that the participant perceived the emotional intent as ambiguous. As in Appendix C, we estimated mixed-effects logistic models with random intercepts for participants and emotion categories.

| Response Option | Count |
|---|---|
| Happy (correct) | 196 |
| Amused | 67 |
| Disappointment | 4 |
| Nervous | 62 |
| No emotion | 72 |
| Unsure / Other | 88 |
| Total | 489 |

Table 8: Response distribution for Example 2

| Comparison | $t$ | df | Mean diff. | 95% CI |
|---|---|---|---|---|
| Non-sarcastic: non-eNVC vs. eNVC | −17.18 | 512 | −1.32 | [−1.47, −1.17] |
| Sarcastic: eNVC vs. non-eNVC | −12.17 | 512 | −0.92 | [−1.07, −0.77] |

Table 9: Paired-samples *t*-tests comparing accuracy across eNVC conditions.

## Model 1: Main Effects of eNVCs and Sarcasm

Model 1 (Table 13) assessed the independent effects of eNVCs and sarcasm on the likelihood of choosing "Uncertain." The effect of eNVCs was small and not significant ($\hat{\beta}$ = −0.015, SE = 0.070, $p$ = .832), indicating that preserving nonverbal cues did not reliably reduce perceived ambiguity. By contrast, sarcasm was a strong positive predictor of uncertainty ($\hat{\beta}$ = 0.990, SE = 0.062, $p$ < .001, OR = 2.69), meaning sarcastic posts were nearly three times as likely to be labeled "Uncertain" as literal ones. White participants showed higher uncertainty rates than others, but other demographic effects were near zero.

## Model 2: Four-Condition Specification

Model 2 (Table 14) decomposed eNVC × sarcasm combinations into four mutually exclusive conditions using the *non-eNVC, non-sarcastic* cell as the reference. Estimated marginal means are reported in Table 15.

Relative to baseline:

- **eNVC, non-sarcastic** posts showed no meaningful change in uncertainty ($\hat{\beta}$ = −0.024, SE = 0.071, $p$ = .732).

- **Non-eNVC, sarcastic** posts significantly increased uncertainty ($\hat{\beta}$ = 0.743, SE = 0.352, $p$ = .035, OR = 2.10).

- **eNVC, sarcastic** posts produced the strongest increase in uncertainty ($\hat{\beta}$ = 0.973, SE = 0.063, $p$ < .001, OR = 2.65).

This pattern mirrors the accuracy results in Appendix C: sarcasm drives large increases in perceived ambiguity, and eNVCs do not attenuate this effect. Among sarcastic posts, preserving cues does not decrease uncertainty relative to removing them.

Across both uncertainty models, sarcasm is the dominant predictor of perceived ambiguity, doubling or tripling the likelihood that participants select the "Uncertain" option. In contrast, eNVCs do not reliably reduce uncertainty and, in sarcasm conditions, do not offset the interpretive difficulty introduced by ironic reversal. Together with accuracy analyses, these results highlight a boundary condition on the communicative value of nonverbal cues in text: eNVCs aid literal decoding but do not mitigate the uncertainty inherent in sarcastic communication.

| Predictor | Estimate | SE | z | p |
|---|---|---|---|---|
| Intercept | 0.390 | 0.236 | 1.65 | .099 |
| eNVC (present) | 0.559 | 0.047 | 11.79 | < .001 |
| Sarcasm (yes) | -1.274 | 0.048 | -26.51 | < .001 |
| Gender (male) | -0.010 | 0.040 | -0.24 | .808 |
| White | 0.171 | 0.041 | 4.14 | < .001 |
| Age | -0.022 | 0.018 | -1.22 | .221 |
| Random effects: | Participant SD = 0.256; Emotion SD = 0.562 | | | |
| N observations: | 16,352 (511 participants, 6 emotions) | | | |

Table 10: Mixed-effects logistic regression (Model 1): Main effects of eNVCs and sarcasm.

| Predictor | Estimate | SE | z | p |
|---|---|---|---|---|
| Intercept | 0.204 | 0.181 | 1.13 | .260 |
| eNVC, non-sarcastic | 0.565 | 0.047 | 11.92 | < .001 |
| Non-eNVC, sarcastic | -0.171 | 0.422 | -0.41 | .685 |
| eNVC, sarcastic | -0.715 | 0.046 | -15.43 | < .001 |
| Gender (male) | -0.010 | 0.040 | -0.24 | .808 |
| White | 0.171 | 0.041 | 4.14 | < .001 |
| Age | -0.022 | 0.018 | -1.22 | .221 |
| Random effects: | Participant SD = 0.256; Emotion SD = 0.384 | | | |
| N observations: | 16,352 (511 participants, 6 emotions) | | | |

Table 11: Mixed-effects logistic regression (Model 2): Four-condition specification.

Table 12: Estimated marginal means (EMMs) for accuracy by condition (Study 2). Higher values indicate greater probability of correct emotion identification.

| Condition | Pred. Prob. | 95% CI | SE |
|---|---|---|---|
| Non-eNVC, Non-sarcastic | 0.56 | [0.47, 0.64] | 0.043 |
| eNVC, Non-sarcastic | 0.69 | [0.61, 0.76] | 0.038 |
| Non-eNVC, Sarcastic | 0.51 | [0.33, 0.69] | 0.096 |
| eNVC, Sarcastic | 0.38 | [0.30, 0.46] | 0.041 |

| Predictor | Estimate | SE | z | p |
|---|---|---|---|---|
| Intercept | -2.612 | 0.173 | -15.13 | < .001 |
| eNVC (present) | -0.015 | 0.070 | -0.21 | .832 |
| Sarcasm (yes) | 0.990 | 0.062 | 15.86 | < .001 |
| Gender (male) | -0.044 | 0.078 | -0.57 | .572 |
| White | 0.563 | 0.079 | 7.12 | < .001 |
| Age | 0.016 | 0.034 | 0.47 | .635 |
| Random effects: | Participant SD = 0.695; Emotion SD = 0.330 | | | |
| N observations: | 16,352 (511 participants, 6 emotions) | | | |

Table 13: Mixed-effects logistic regression (Model 1): Main effects on perceived uncertainty.

| Predictor | Estimate | SE | z | p |
|---|---|---|---|---|
| Intercept | -2.567 | 0.180 | -14.29 | < .001 |
| eNVC, non-sarcastic | -0.024 | 0.071 | -0.34 | .732 |
| Non-eNVC, sarcastic | 0.743 | 0.352 | 2.11 | .035 |
| eNVC, sarcastic | 0.973 | 0.063 | 15.47 | < .001 |
| Gender (male) | -0.044 | 0.078 | -0.57 | .572 |
| White | 0.563 | 0.079 | 7.12 | < .001 |
| Age | 0.016 | 0.034 | 0.47 | .635 |
| Random effects: | Participant SD = 0.695; Emotion SD = 0.317 | | | |
| N observations: | 16,352 (511 participants, 6 emotions) | | | |

Table 14: Mixed-effects logistic regression (Model 2): Four-condition specification for uncertainty.

| Condition | Pred. Prob. | 95% CI | SE |
|---|---|---|---|
| Non-eNVC, Non-sarcastic | 0.094 | [0.071, 0.123] | 0.013 |
| eNVC, Non-sarcastic | 0.092 | [0.070, 0.120] | 0.013 |
| Non-eNVC, Sarcastic | 0.179 | [0.104, 0.290] | 0.047 |
| eNVC, Sarcastic | 0.215 | [0.170, 0.269] | 0.025 |

Table 15: Estimated marginal means (EMMs) for uncertainty by condition (Study 2). Higher values indicate greater perceived ambiguity.

# Appendix E: Study 3 Screening Survey stimuli

Table 16 reports the selection of trending topics and hashtags used for selecting the stimuli for the screening survey of Study 3.

# Appendix F: Age of Participants in Study 3 Screening Survey

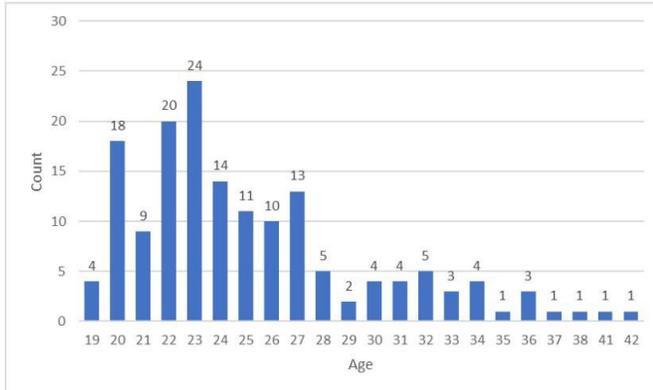

Figure 3: Bar graph of age distribution of participants in Study 3 screening survey.

As shown in Figure 3, the majority of participants in the Study 3 screening survey were between 18 and 30 years old, with representation tapering off in older cohorts. This skew toward younger demographics is expected given the participant pool, but the inclusion of participants across age ranges allowed for diversity in interpretations of electronic nonverbal cues (eNVCs). The age spread ensured that while Gen Z voices were dominant, perspectives from Millennials, Gen X, and Baby Boomers were also included, strengthening the robustness of qualitative comparisons.

| No. | Trending Section | Topic and Search Query | Description |
|---|---|---|---|
| 1 | Trending | "#TheLastOfUs" (#TheLastOfUs) lang:en -filter:links -filter:replies | The Last of Us, a popular video game series, trended as HBO released the trailer for its TV adaptation. |
| 2 | Trending | "Twitter's TOS" Twitter's TOS lang:en -filter:links -filter:replies | Changes were recently made to Twitter's terms of service. |
| 3 | Gaming - Trending | "#TheCallistoProtocol" (#TheCallistoProtocol) lang:en -filter:links -filter:replies | Hashtag for a new videogame developed by Striking Distance Studios, released on Dec 2, 2022. |
| 4 | Music - Trending | "Wonho" Wonho lang:en -filter:links -filter:replies | Wonho, South Korean singer from Monsta X, held his last concert before entering compulsory military service. Fans trended his name to express support. |
| 5 | Trending | "Kanye" Kanye lang:en -filter:links -filter:replies | Kanye West trended after suspension from Twitter for antisemitic posts and a controversial image. Some users speculated suspension was retaliation for posting an unflattering image of Elon Musk. |
| 6 | Gaming - Trending | "#DragonAge" (#DragonAge) lang:en -filter:links -filter:replies | Trending during the annual Dragon Age community celebration on Dec 4. |
| 7 | NFL - Trending | "Colts" Colts lang:en -filter:links -filter:replies | Trended due to an NFL game featuring the Colts. |
| 8 | Entertainment - Trending | "The Way of Water" "The Way of Water" lang:en -filter:links -filter:replies | Trending due to anticipation for James Cameron's second Avatar movie. |
| 9 | Trending | "Portugal" "Portugal" lang:en -filter:links -filter:replies | Trended during a sports event featuring the Portuguese national team. |
| 10 | Sports - Trending Trending with #NBAonTNT | "Shaq" "Shaq" lang:en -filter:links -filter:replies | Trended during a sports broadcast featuring Shaquille O'Neal. |

Table 16: Trending topics and hashtags sampled for screening survey (Study 3).

# Appendix G: Focus Group Discussion Protocol

**Expected duration of focus group discussion:** around 45 minutes to 1 hour

**Type of focus group discussion:** Semi-structured

**Research question to be answered:** (RQ3) How do people decode emotions using nonverbal cues in microblog posts?

| No. | Question | Objective |
| --- | --- | --- |
| 1 | Please read this and think about it. In your own words, what do you think this tweet is about? | To get participants to start thinking about the meaning of the microblog post. Participants may identify the subject, overall tone, emotions, eNVCs, and audiences amongst other details from the post. |
| 2 | What are some emotions you can identify in this tweet? *Feel free to be as specific as you'd like. You are also encouraged to work off each others' responses if you agree or disagree with their response.* | To get participants to think deeply about the specific emotions they see in the post. |
| 3 | Now, what are some of the ways you think the author expresses these emotions? *Think about anything in the tweet that signals certain emotions in you while reading.* | To get participants to identify details in the post that helped them identify the emotions they had previously named. This also helps tie in the emotions that participants may have felt the post embodies, with the actual written content of the post. |

# Appendix H: Focus Group Sessions and Posts Shown

| No. | Post | S1 | S2 | S3 | S4 | S5 | S6 |
| --- | --- | --- | --- | --- | --- | --- | --- |
| 1 | Today is Friday!!!!!!! thannkkkk gooood | ✓ | | | ✓ | | |
| 2 | i fucking told all of them (my friends) to save a seat for me... sigh | ✓ | ✓ | ✓ | | | |
| 3 | I have so much homework & I'm overthinking.. 😔 gawd I'm such a mess... | ✓ | | | ✓ | | |
| 4 | I'm shit at everything I do 👍 | ✓ | | | ✓ | | |
| 5 | Just finished reading Farenheit 451... DAMN... I really enjoyed it 👌 | ✓ | | ✓ | | | |
| 6 | IM CUTTING MY HAIR TODAy!!!!!!! | | ✓ | ✓ | | | |
| 7 | all my friends hate me... haha kill me | | ✓ | ✓ | | | |
| 8 | Oh I am soo tiiiired ughh wow I am so tired... | | | | ✓ | ✓ | ✓ |
| 9 | ssooooo i see a coffee machine here... oh myggod | | | | | ✓ | ✓ |
| 10 | I think my record player just came in the mail 😍😍😍 THANK YOU FEDEX | | | | | ✓ | ✓ |
| 11 | EXAAAAAAAAAMS 😨 | | | | | ✓ | ✓ |
| 12 | When the guy you like starts talking to another girl.. 👌 | | | | | ✓ | ✓ |

Table 18: Posts and focus group sessions in Study 3. A checkmark (✓) indicates that the post was discussed in the given session.